\documentclass[lettersize,journal]{IEEEtran}
\usepackage{amsmath,amsfonts}
\usepackage{algorithmic}
\usepackage{algorithm}
\usepackage{array}
\usepackage[caption=false,font=normalsize,labelfont=sf,textfont=sf]{subfig}
\usepackage{textcomp}
\usepackage{stfloats}
\usepackage{url}
\usepackage{verbatim}
\usepackage{graphicx}
\usepackage{multirow}
\usepackage{cite}
\usepackage{colortbl}  
\usepackage{xcolor}
\usepackage{adjustbox}
\usepackage{tabularx}
\usepackage{pifont} 
\usepackage{booktabs}
\usepackage[normalem]{ulem}

\begin{document}

\title{FoCal: Frequency-Oriented Cross-Modal Interaction and Spectral Calibration for Aerial Visible-Infrared Object Detection}

\author{Ben Liang,~\IEEEmembership{Graduate Student Member,~IEEE}, Chao Sui, Junqi Bai, Yuan Liu,~\IEEEmembership{Member,~IEEE}, Chunlai Li, Xiubao Sui, and Qian Chen
\thanks{This work was supported in part by the National Natural Science Foundation
of China under Grant 62427818, Grant U25A20502, and Grant 62301253, and in part by the Fundamental Research Funds for the Central Universities under Grant 30925010511. (Ben Liang and Chao Sui contributed equally to this work.) (Corresponding authors: Yuan Liu and Xiubao Sui.)}

\thanks{Ben Liang, Chao Sui, Yuan Liu, Xiubao Sui, and Qian Chen are with the School of Electronic and Optical Engineering, Nanjing University of Science and Technology, Nanjing 210094, China (e-mail: benliang@njust.edu.cn; suichao@njust.edu.cn; lyuan90\_eo@njust.edu.cn; sxb@njust.edu.cn; chenqian@njust.edu.cn).}

\thanks{Chunlai Li is with the School of Electronic Science and Technology, Shanghai Institute of Technical Physics, Chinese Academy of Sciences, Shanghai 200083, China (e-mail: lichunlai@mail.sitp.ac.cn).}

\thanks{Junqi Bai is with the 28th Institute of China Electronics Technology Group Corporation, Nanjing 210007, China (baijunqi@cetc.com.cn).}

\thanks{Xiubao Sui is also with the State Key Laboratory of Extreme Environment Optoelectronic Dynamic Measurement Technology and Instrument, Nanjing University of Science and Technology, Nanjing 210094, China.}
}
\markboth{Journal of \LaTeX\ Class Files,~Vol.~14, No.~8, August~2021}%
{Shell \MakeLowercase{\textit{et al.}}: A Sample Article Using IEEEtran.cls for IEEE Journals}


\maketitle

\begin{abstract}
In aerial RGB--IR object detection, effectively exploiting complementary information across modalities is critical for robust perception under complex illumination and environmental conditions. Existing multimodal detectors mainly focus on spatial-domain interaction or frequency-specific feature enhancement, while the cross-modal interaction patterns of different frequency components remain insufficiently explored. Moreover, spectral discrepancy itself may contain both useful complementary cues and unreliable modality-specific responses, making indiscriminate frequency fusion suboptimal. To address these issues, we propose FoCal, a frequency-oriented framework for aerial RGB--IR object detection. First, a Frequency-Aware Dual-Domain Calibration (FADC) module is developed to explicitly model frequency-dependent cross-modal interaction. Low-frequency components are collaboratively consolidated into a shared structural consensus, whereas high-frequency components preserve modality-specific information through selective cross-modal exchange. The resulting frequency-aware cues are further transferred to the original feature domain to regulate cross-modal calibration. Second, we introduce a Discrepancy-Guided Spectral Modulation (DGSM) module, which characterizes cross-modal spectral imbalance using confidence-weighted relative amplitude discrepancy and transforms it into a bounded signed gate for adaptive enhancement, preservation, or attenuation of the joint multimodal spectrum. Extensive experiments on DroneVehicle, ESCVehicle, and ATR-UMOD demonstrate the effectiveness of FoCal, yielding $\mathrm{mAP}_{50}$ values of 83.5\%, 54.8\%, and 64.6\%, respectively. Meanwhile, with only 3.0M parameters, FoCal achieves 113.6 FPS while preserving leading detection accuracy, highlighting a favorable accuracy--efficiency trade-off. Code is available at \url{https://github.com/universeliang/FoCal.}
\end{abstract}

\begin{IEEEkeywords}
object detection, multimodal, cross-modal interaction, frequency fusion, modality-specific.
\end{IEEEkeywords}

\section{Introduction}
In recent years, single-modality object detection based on visible (RGB) imagery has achieved remarkable progress and has been widely applied in various scenarios, including intelligent transportation \cite{liu2026d2,liang2026multi}, disaster rescue \cite{wang2025efficient,liu2025dmsa}, and remote sensing \cite{wei2024spatio,wei2025multi,chen2026tiny}. Nevertheless, RGB imaging is highly dependent on external illumination, and its visual quality can deteriorate substantially under low illumination, strong light interference, and adverse weather conditions, leading to severe detail loss and contrast degradation \cite{sun2022drone,shi2025spgfuse}. In contrast, infrared (IR) sensors capture the thermal radiation emitted by objects and are therefore less sensitive to illumination variations, while also providing complementary object contours and structural cues under challenging environments \cite{jiang2024cross,li2025pseudo}. These complementary imaging characteristics make RGB--IR multimodal fusion an effective paradigm for robust all-weather visual perception \cite{tang2023camf,li2024m2fnet,shen2024icafusion}.

\begin{figure}[!t]
    \centering
    \includegraphics[width=\columnwidth]{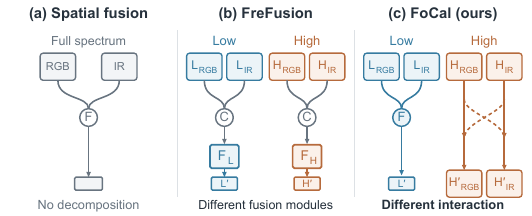}
    \caption{Comparison of cross-modal fusion paradigms. (a) Spatial-domain fusion directly combines RGB and IR features. (b) FreFusion combines same-frequency components before applying band-specific fusion modules. (c) FoCal differentiates the interaction itself: low-frequency components form a shared structural consensus, whereas high-frequency components undergo selective exchange while preserving modality-specific details.}
    \label{fig-focal_core_compact}
\end{figure}

\begin{figure}[!t]
    \centering
    \includegraphics[width=.9\columnwidth]{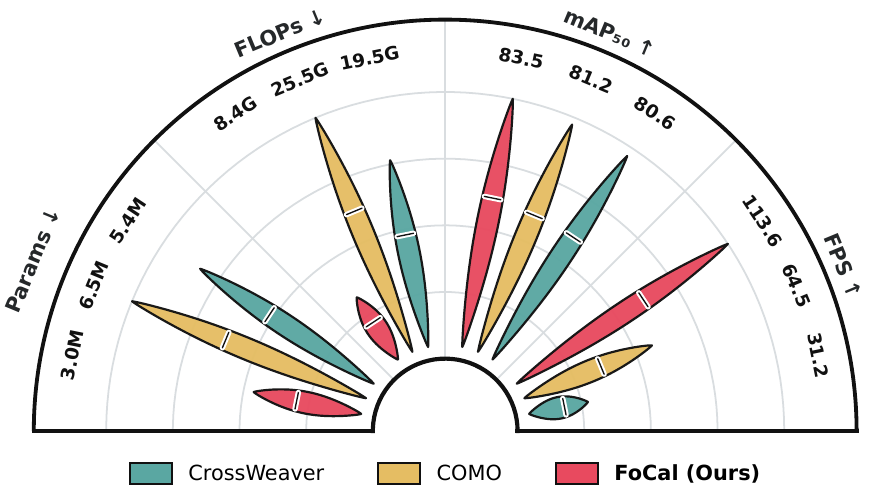}
    \caption{Comparison of model accuracy and computational efficiency. Petal lengths are linearly scaled to the original values within each metric, with the maximum value assigned the longest petal. FoCal achieves the highest $\mathrm{mAP}_{50}$ and FPS while requiring the fewest parameters and FLOPs.}
    \label{fig-acc-fps}
\end{figure}

Multimodal fusion detection (MFD) aims to exploit both shared and modality-specific information from heterogeneous sensors to construct robust representations for object detection \cite{chen2023igt}. Most existing approaches perform cross-modal interaction primarily in the spatial domain, where semantic alignment, feature enhancement, or attention-based aggregation is employed to improve multimodal complementarity \cite{shen2024icafusion,zhang2023superyolo}. However, spatial-domain interaction alone does not explicitly distinguish structural information from fine-grained modality-dependent responses. Recent studies have therefore introduced frequency-domain decomposition to separate low-frequency structures from high-frequency details and assign different enhancement operators to individual frequency bands \cite{li2025fd2,zhu2025wavemamba}. Despite their effectiveness, these approaches mainly focus on how different frequency components are processed after decomposition, while a more fundamental question remains underexplored: should cross-modal components at different frequencies follow the same interaction pattern in the first place?

As shown in Fig.~\ref{fig-focal_core_compact}, we argue that frequency characteristics should determine not only the subsequent processing operators, but also the cross-modal information-flow topology. Low-frequency components mainly encode coarse contours, object layouts, and large-scale structures, which generally exhibit stronger correspondence between RGB and IR modalities. They are therefore more suitable for consensus formation, where both modalities collaboratively establish a shared structural representation. High-frequency components, in contrast, contain fine edges and local textures together with modality-dependent responses and sensor-specific noise. Prematurely collapsing them into a single representation may suppress useful modality-specific details or propagate unreliable responses across modalities. From this perspective, low- and high-frequency components require fundamentally different cross-modal interaction patterns: a \emph{many-to-one consensus} for low-frequency structures and an \emph{individuality-preserving two-to-two exchange} for high-frequency details. More importantly, frequency-domain interaction should not be regarded as the endpoint of multimodal fusion. We consider the frequency domain as an auxiliary space for estimating the reliability of cross-modal interaction. The resulting frequency-conditioned cues are transferred back to the original feature domain to regulate information injection between RGB and IR streams. In this manner, frequency-specific interaction and original-domain feature calibration are explicitly coupled while serving different purposes: the former determines how the modalities should interact, whereas the latter determines where and to what extent counterpart information should be introduced.

Based on these observations, we propose FoCal, a frequency-oriented framework for aerial RGB--IR object detection. First, we propose a Frequency-Aware Dual-Domain Calibration (FADC) module that explicitly models frequency-dependent cross-modal interaction. It promotes structural consensus in low-frequency components while preserving modality individuality through selective high-frequency exchange, and further transfers the learned frequency cues to calibrate the original feature streams. Second, we propose a Discrepancy-Guided Spectral Modulation (DGSM) module that exploits confidence-weighted spectral discrepancy as a control signal to adaptively regulate the joint multimodal spectrum. It converts the learned discrepancy cues into a signed spectral gate, enabling frequency-wise enhancement, preservation, or attenuation of the joint representation. Extensive experiments are conducted on three aerial-view multimodal datasets, including DroneVehicle \cite{sun2022drone}, ESCVehicle \cite{song2026escvehicle}, and ATR-UMOD \cite{chen2025fusion}. FoCal achieves $\mathrm{mAP}_{50}$ values of $83.5\%$, $54.8\%$, and $64.6\%$ on the three datasets, respectively, consistently delivering state-of-the-art performance across all three benchmarks. As illustrated in Fig.~\ref{fig-acc-fps}, FoCal further achieves 113.6 FPS with only 3.0M parameters and 8.4G FLOPs, highlighting its favorable accuracy--efficiency trade-off. Overall, our main contributions are summarized as follows:

(1) We propose FoCal, a lightweight frequency-oriented framework for aerial RGB--IR object detection. Rather than merely assigning different processing operators to different frequency bands, FoCal explicitly models frequency-dependent cross-modal interaction and further exploits spectral discrepancy to regulate multimodal representation learning.

(2) We propose a Frequency-Aware Dual-Domain Calibration (FADC) module that differentiates the cross-modal interaction patterns of low- and high-frequency components. Low-frequency representations are collaboratively consolidated to establish structural consensus, whereas high-frequency representations preserve modality individuality through selective cross-modal exchange. The resulting frequency-aware cues are further transferred to the original feature domain for cross-modal calibration.

(3) We propose a Discrepancy-Guided Spectral Modulation (DGSM) module that characterizes cross-modal spectral imbalance using confidence-weighted relative amplitude discrepancy. Instead of directly treating discrepancy as informative content, DGSM converts it into a bounded signed control signal to adaptively enhance, preserve, or attenuate the joint multimodal spectrum.
\section{Related Work}
\label{sec2:Related Work}
\subsection{Multimodal Object Detection}
\label{sec2.1}
Multimodal fusion aims to exploit complementary information from different modalities to produce high-quality fused representations \cite{shi2025infrared}, thereby improving the performance of MFD. For example, CDDFuse \cite{zhao2023cddfuse} and MambaDFuse \cite{li2024mambadfuse} employ Transformer- and Mamba-based architectures, respectively, to enhance pixel-level fusion through global contextual modeling, and subsequently transfer the fused representations to downstream detection tasks. Owing to the complementary properties of different modalities, there has been growing interest in using RGB--IR image pairs for multispectral pedestrian and vehicle detection. UA-CMDet \cite{sun2022drone} introduces DroneVehicle, an RGB--IR vehicle detection benchmark captured from UAV viewpoints, and improves detection performance by leveraging illumination estimation to model cross-modal uncertainty. CFT \cite{qingyun2021cross} and YOLOFusion \cite{qingyun2022cross} enhance multispectral object detection by modeling global feature interactions both across and within modalities. ICAFusion \cite{shen2024icafusion} proposes a query-guided iterative cross-attention mechanism to capture cross-modal complementary information and improve feature discriminability. FusionMamba \cite{dong2025fusion} further projects cross-modal features into a hidden state space via Mamba for enhanced representation learning, thereby suppressing pseudo-target responses. Although these methods have achieved promising performance, there remains substantial room for improvement. In particular, frequency-domain disentanglement provides a promising route to separate informative components for more refined cross-modal fusion, which is especially beneficial for small-object detection in complex aerial scenarios.

\subsection{Frequency in computer vision}
\label{sec2.2}
Frequency-domain analysis has been increasingly explored in recent vision tasks due to its ability to disentangle low- and high-frequency components for more effective representation learning \cite{zhang2025sfnet,shi2025hs}. Among representative frequency-domain techniques, Fast Fourier Transform (FFT) and Discrete Wavelet Transform (DWT) exhibit complementary characteristics: FFT is particularly effective at modeling global frequency interactions, whereas DWT is better suited for local multi-scale decomposition. DFFormer \cite{xiang2025dfformer} replaces conventional MHSA with an FFT-based token mixer to enhance global modeling, achieving excellent performance on high-resolution images with lower computational cost. SFDFusion \cite{hu2024sfdfusion} employs FFT to fuse the amplitude and phase of infrared and visible images, and further cascades the spectral representations with spatial-domain features, thereby effectively improving multimodal image fusion quality. WTConv \cite{finder2024wavelet} effectively enlarges the effective receptive field through cascaded wavelet transforms. HaarFuse \cite{wang2025haarfuse} decomposes features into high- and low-frequency subbands using wavelet transform, and then applies CNNs and Transformers to finely refine different components, resulting in improved fusion quality. In multimodal detection, WaveMamba \cite{zhu2025wavemamba} efficiently combines wavelet decomposition with Vision Mamba, enhancing the capability of multimodal object detection. DEPFusion \cite{li2025depfusion} leverages Mamba and FFT to enhance multi-scale context and texture details in RGB low-frequency components, thereby improving UAV multispectral object detection performance. Despite recent progress, most existing multimodal detection methods still assume equal contributions of frequency components from different modalities, while overlooking the trade-off relationships between same-frequency components across modalities. Consequently, frequency-dependent cross-modal interaction and discrepancy-aware spectral modeling remain insufficiently explored for refined multimodal fusion. Moreover, these methods usually treat frequency-enhanced features as the final fusion outputs, without further exploiting them to regulate the subsequent interaction of the original feature streams.
\section{Methods}
\label{sec3:section}
\begin{figure*}[!t]
  \includegraphics[width=\textwidth]{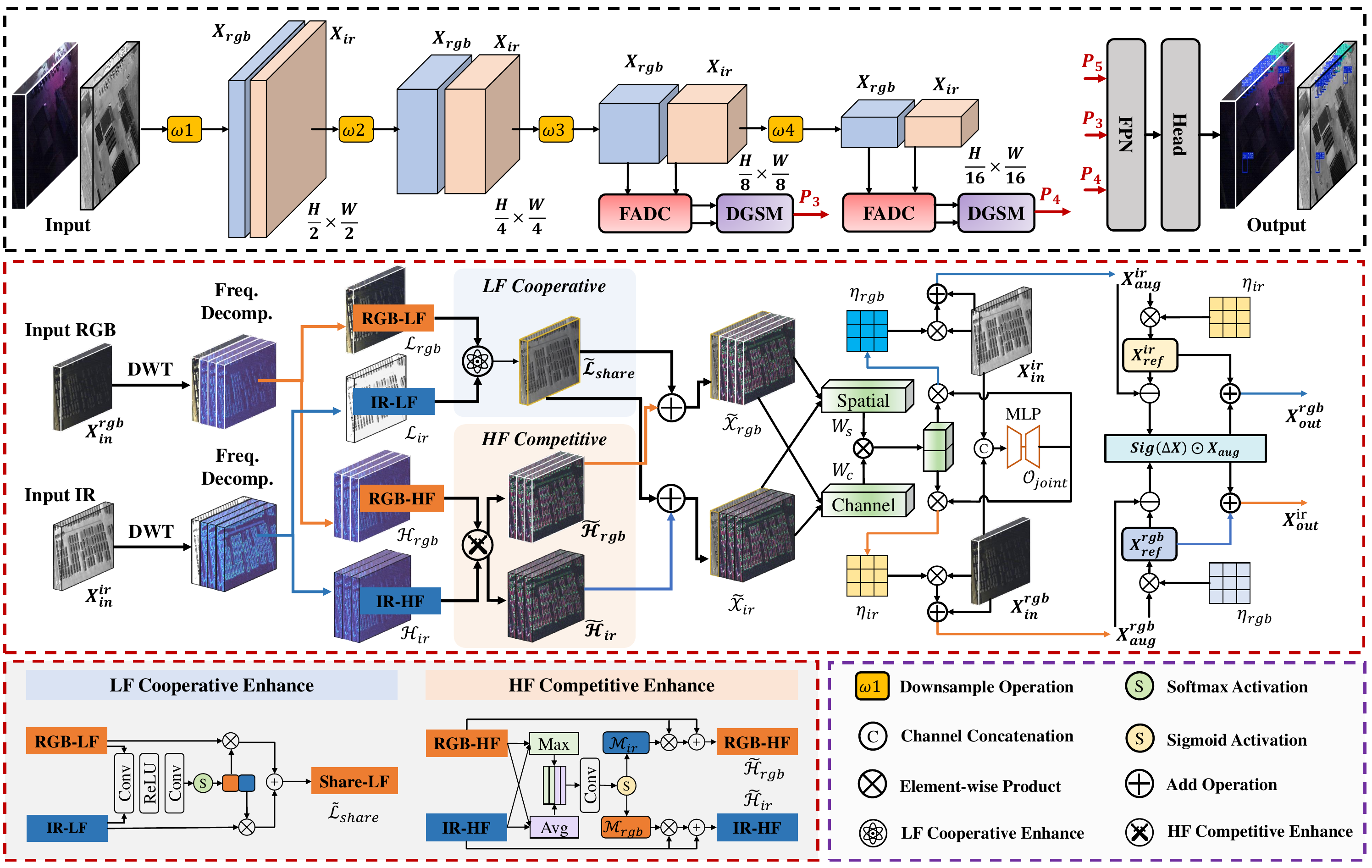}
  \caption{Overall architecture of the proposed FoCal. The upper part presents the macroscopic detection pipeline, which consists of a Backbone for feature extraction, a Path Aggregation Network (PAN) for multi-scale feature fusion, and a Detection Head.}
  \label{fig-FoCal}
\end{figure*}

In this section, we first present the overall architecture of the proposed FoCal network, as shown in Fig.~\ref{fig-FoCal}. We then elaborate on the design rationale of the FADC and DGSM modules, respectively.

\subsection{Frequency-Aware Dual-Domain Calibration}
\label{sec3.1:subsection}
RGB and infrared sensors exhibit substantially different imaging characteristics, resulting in distinct cross-modal behaviors across frequency bands. Existing frequency-aware fusion methods typically distinguish low- and high-frequency components by assigning them different enhancement operators after decomposition. However, such a strategy mainly focuses on how frequency components are processed after interaction, while paying less attention to a more fundamental issue: whether different frequency bands should follow the same cross-modal interaction pattern in the first place. We argue that frequency specificity should determine not only the subsequent processing operator, but also the cross-modal information-flow topology. In particular, low-frequency components mainly encode coarse structures and semantic layouts that exhibit relatively strong cross-modal correspondence, and are therefore suitable for establishing a shared structural consensus. In contrast, high-frequency components contain modality-dependent contours, textures, and local details, together with sensor-specific noise. Prematurely collapsing them into a single fused representation may suppress useful modality-specific cues or propagate unreliable responses across modalities. 

Motivated by this observation, we propose a Frequency-Aware Dual-Domain Calibration (FADC) module. FADC explicitly assigns different interaction topologies to different frequency bands: low-frequency representations are collaboratively consolidated into a shared structural consensus, whereas high-frequency representations remain modality-specific and interact through recipient-conditioned selective exchange. The resulting frequency-conditioned representations are further used to estimate cross-modal reliability and calibrate information transfer over the original feature streams.

Given the feature maps $X_{rgb}, X_{ir} \in \mathbb{R}^{C \times H \times W}$ from the RGB and IR branches, we first employ a Discrete Wavelet Transform (DWT) with the Haar basis to separate coarse structures from directional details. Specifically, each input feature is decomposed into one low-frequency approximation component and three directional high-frequency components:
\begin{equation}
\{\mathcal{A}, \mathcal{H}^{h}, \mathcal{H}^{v}, \mathcal{H}^{d}\}
=
\mathcal{W}(X),
\label{eq:haar_decompose}
\end{equation}
where $\mathcal{W}(\cdot)$ denotes the Haar-based DWT, $\mathcal{A}$ represents the low-frequency approximation, and $\mathcal{H}^{h}$, $\mathcal{H}^{v}$, and $\mathcal{H}^{d}$ denote horizontal, vertical, and diagonal high-frequency responses, respectively. For each modality, the directional high-frequency components are aggregated as
\begin{equation}
\mathcal{H}_{m}
=
\sum_{k \in \{h,v,d\}}
\mathcal{H}_{m}^{k},
\qquad
m \in \{rgb, ir\}.
\label{eq:overall_decouple}
\end{equation}
Accordingly, each modality is represented by a low-frequency structural component $\mathcal{A}_{m}$ and a high-frequency detail component $\mathcal{H}_{m}$.

\subsubsection{\textbf{Frequency-Dependent Interaction Topology}}
A key design principle of FADC is that low- and high-frequency components should not simply adopt different post-fusion operators; instead, they should follow different cross-modal interaction topologies according to their information characteristics.

\emph{Low-Frequency Consensus Formation.}
Low-frequency responses primarily describe object layouts, coarse contours, and large-scale semantic structures, which generally exhibit stronger cross-modal correspondence than fine-grained details. Therefore, instead of independently enhancing $\mathcal{A}_{rgb}$ and $\mathcal{A}_{ir}$ after direct aggregation, we explicitly establish a shared structural anchor before subsequent feature calibration. Specifically, the two low-frequency components are concatenated and fed into a lightweight convolutional gating function followed by Softmax normalization to obtain complementary weights $\eta_{rgb}$ and $\eta_{ir}$. The shared low-frequency consensus is formulated as
\begin{equation}
\mathcal{A}_{shared}
=
\eta_{rgb}\odot\mathcal{A}_{rgb}
+
\eta_{ir}\odot\mathcal{A}_{ir},
\quad
\eta_{rgb}+\eta_{ir}=1.
\label{eq:lf_consensus}
\end{equation}
This operation implements a many-to-one interaction topology,
\begin{equation}
(\mathcal{A}_{rgb},\mathcal{A}_{ir})
\rightarrow
\mathcal{A}_{shared},
\end{equation}
where both modalities collaboratively determine a common structural representation. The adaptive weighting prevents the shared representation from degenerating into a fixed average and allows the network to dynamically adjust the contribution of each modality according to local observation quality.

\emph{High-Frequency Selective Exchange.}
High-frequency components exhibit a fundamentally different cross-modal relationship. They contain useful modality-specific edges and textures, but also sensor-dependent noise and local disturbances. Directly collapsing $\mathcal{H}_{rgb}$ and $\mathcal{H}_{ir}$ into a single high-frequency representation may therefore discard modality-specific details or allow unreliable responses from one modality to dominate the other. Instead of forming a shared high-frequency representation, FADC preserves both modality-specific streams and performs \emph{recipient-conditioned selective exchange}. We first derive channel-aggregated spatial descriptors using channel-wise average and max pooling. The descriptors from the two modalities are jointly encoded by a lightweight convolution-Sigmoid gating function to generate two modality-specific reception masks, $\mathcal{M}_{rgb}$ and $\mathcal{M}_{ir}$. The high-frequency components are then updated as
\begin{equation}
\tilde{\mathcal{H}}_{rgb}
=
\mathcal{H}_{rgb}
+
\mathcal{M}_{rgb}\odot\mathcal{H}_{ir},
\quad
\tilde{\mathcal{H}}_{ir}
=
\mathcal{H}_{ir}
+
\mathcal{M}_{ir}\odot\mathcal{H}_{rgb}.
\label{eq:hf_exchange}
\end{equation}
Unlike winner-take-all selection or direct high-frequency aggregation, Eq.~\eqref{eq:hf_exchange} follows a two-to-two interaction topology,
\begin{equation}
(\mathcal{H}_{rgb},\mathcal{H}_{ir})
\rightarrow
(\tilde{\mathcal{H}}_{rgb},\tilde{\mathcal{H}}_{ir}),
\end{equation}
where each modality preserves its intrinsic high-frequency responses while selectively absorbing complementary details from its counterpart. Importantly, $\mathcal{M}_{rgb}$ and $\mathcal{M}_{ir}$ characterize the \emph{reception preference} of the corresponding target modalities rather than the standalone saliency of the source modality. This recipient-conditioned interaction prevents premature modality collapse and reduces the propagation of modality-specific noise.

The low- and high-frequency representations are subsequently recomposed into two frequency-conditioned features:
\begin{equation}
\tilde{X}_{rgb}
=
\mathcal{A}_{shared}
+
\tilde{\mathcal{H}}_{rgb},
\quad
\tilde{X}_{ir}
=
\mathcal{A}_{shared}
+
\tilde{\mathcal{H}}_{ir}.
\label{eq:freq_recompose}
\end{equation}
Therefore, the two modalities share the same low-frequency structural anchor while retaining individually calibrated high-frequency details.

\subsubsection{\textbf{Frequency-Guided Reliability Calibration}}
The frequency-conditioned features in Eq.~\eqref{eq:freq_recompose} are not directly regarded as the final fused representation. Instead, they serve as auxiliary cues for estimating where and what information can be reliably transferred between the two original modality streams. Specifically, we derive spatial and channel reliability masks from $\tilde{X}_{rgb}$ and $\tilde{X}_{ir}$:
\begin{equation}
W_s
=
\mathcal{S}(\tilde{X}_{rgb},\tilde{X}_{ir}),
\quad
W_c
=
\mathcal{C}(\tilde{X}_{rgb},\tilde{X}_{ir}),
\label{eq:reliability}
\end{equation}
\begin{equation}
W
=
W_s\odot W_c,
\label{eq:reliability2}
\end{equation}
where $W=\{W_{rgb},W_{ir}\}$. The spatial weighting function $\mathcal{S}(\cdot)$ extracts position-sensitive reliability from channel-wise average and maximum responses, whereas the channel weighting function $\mathcal{C}(\cdot)$ models channel dependencies from pooled bimodal statistics through a shared MLP. Their combination provides modality-specific frequency-conditioned reliability estimates. Meanwhile, the original RGB and IR features are jointly encoded to obtain a bimodal contextual prior:
\begin{equation}
\mathcal{O}_{joint}
=
\mathcal{G}([X_{rgb},X_{ir}]),
\label{eq:joint_prior}
\end{equation}
where $\mathcal{G}(\cdot)$ denotes a lightweight gating function. We then combine the joint bimodal prior with the frequency-conditioned reliability masks to obtain modality-specific calibration masks:
\begin{equation}
\mathcal{R}_{k}
=
\mathcal{O}_{joint}\odot W_{k},
\qquad
k\in\{rgb,ir\}.
\label{eq:calibration_mask}
\end{equation}

Based on $\mathcal{R}_{k}$, complementary information is selectively injected into each original modality stream:
\begin{equation}
X_{k}^{aug}
=
X_{k}
+
\mathcal{R}_{k}\odot X_{\bar{k}},
\qquad
k\in\{rgb,ir\},
\label{eq:cross_injection}
\end{equation}
where $\bar{k}$ denotes the counterpart modality. In this formulation, frequency-domain interaction does not directly replace the original modality representation; instead, it determines the reliability of cross-modal information transfer.

To further regulate the augmented representation, we construct a reliability-conditioned reference feature:
\begin{equation}
X_{k}^{ref}
=
X_{k}^{aug}\odot\mathcal{R}_{k},
\qquad
\Delta X_{k}
=
X_{k}^{ref}-X_{k}^{aug},
\label{eq:reference}
\end{equation}
and perform residual calibration as
\begin{equation}
X_{k}^{out}
=
\sigma(\Delta X_{k})\odot X_{k}^{aug}
+
X_{k}^{ref},
\label{eq:residual_calibration}
\end{equation}
where $\sigma(\cdot)$ denotes the Sigmoid function. This residual calibration further adjusts the contribution of the cross-modally augmented responses according to the learned reliability reference.

Overall, FADC differs from conventional frequency-aware fusion in that frequency decomposition is not merely used to assign different enhancement modules to different frequency bands. Instead, it explicitly imposes frequency-dependent cross-modal interaction topologies: low-frequency components undergo structural consensus formation, whereas high-frequency components preserve modality individuality and perform recipient-conditioned selective exchange. The resulting frequency-conditioned representations are then converted into reliability cues to regulate cross-modal transfer over the original feature streams, thereby avoiding premature modality collapse while retaining complementary information from both modalities.
\subsection{Discrepancy-Guided Spectral Modulation}
\label{sec3.2:dgsm}

Although FADC regulates cross-modal interaction according to frequency characteristics, the resulting modality representations may still exhibit spectral imbalance caused by heterogeneous imaging mechanisms and modality-dependent degradation. Simply aggregating the two modalities may consequently weaken complementary responses or propagate modality-specific interference. To address this issue, we introduce a Discrepancy-Guided Spectral Modulation (DGSM) module. DGSM employs cross-modal spectral discrepancy to guide the modulation of an independently constructed joint spectrum. A confidence-weighted relative discrepancy is first derived from the RGB and IR spectra and then converted into a signed gate for adaptive spectral regulation. The resulting modulation selectively enhances, preserves, or attenuates joint spectral responses according to the learned discrepancy cues, thereby improving spectral complementarity while reducing unreliable modality-specific interference.

Given the modality-specific features $X_{rgb},X_{ir}\in\mathbb{R}^{C\times H\times W}$, we first map them into a common comparison space using a shared projection function $\mathcal{P}(\cdot)$. During training, the RGB and IR features are jointly normalized to reduce modality-dependent feature-scale bias before spectral comparison:
\begin{equation}
X_{rgb}^{p},X_{ir}^{p}
=
\mathcal{P}(X_{rgb},X_{ir}).
\end{equation}
The projected features are then transformed into the frequency domain using the Fast Fourier Transform (FFT):
\begin{equation}
Z_{rgb}=\mathcal{F}(X_{rgb}^{p}),\qquad
Z_{ir}=\mathcal{F}(X_{ir}^{p}),
\end{equation}
where $\mathcal{F}(\cdot)$ denotes the orthonormally normalized FFT. Their amplitude spectra are given by
\begin{equation}
A_{rgb}=|Z_{rgb}|,\qquad
A_{ir}=|Z_{ir}|.
\end{equation}
We therefore define a scale-normalized relative spectral discrepancy as
\begin{equation}
D_{rel}
=
\frac{|A_{rgb}-A_{ir}|}
{A_{rgb}+A_{ir}+\epsilon},
\label{eq:relative_diff}
\end{equation}
where $\epsilon$ is a small constant for numerical stability. In this formulation, $D_{rel}$ characterizes the relative spectral imbalance between the two modalities rather than their absolute energy difference. However, a large relative discrepancy does not necessarily imply useful complementarity. In particular, two nearly inactive spectral responses may still yield a large $D_{rel}$ due to their small denominator. To avoid overemphasizing such unreliable differences, we further introduce an energy confidence term. Let
\begin{equation}
A_{\Sigma}=A_{rgb}+A_{ir},
\end{equation}
and normalize its spectral energy as
\begin{equation}
E=
\frac{A_{\Sigma}}
{\operatorname{Mean}_{\Omega}(A_{\Sigma})+\epsilon},
\end{equation}
where $\operatorname{Mean}_{\Omega}(\cdot)$ denotes averaging over the frequency plane. The corresponding confidence is defined as
\begin{equation}
C_{e}
=
\frac{E}{1+E}.
\end{equation}
The final discrepancy descriptor is therefore
\begin{equation}
D
=
D_{rel}\odot C_{e}.
\label{eq:confidence_diff}
\end{equation}
Consequently, relative differences occurring at nearly inactive spectral locations receive lower confidence, whereas discrepancies supported by meaningful spectral responses are retained for subsequent modulation.

Importantly, DGSM does not directly inject the discrepancy descriptor $D$ into the multimodal representation. Instead, $D$ is transformed into a learnable signed modulation gate:
\begin{equation}
G_{d}
=
\tanh\left(\mathcal{G}_{d}(D)\right),
\qquad
G_{d}\in[-1,1],
\label{eq:signed_gate}
\end{equation}
where $\mathcal{G}_{d}(\cdot)$ denotes a lightweight $1\times1$ convolution. The signed formulation allows the model to learn different responses to cross-modal spectral imbalance. Therefore, spectral discrepancy itself is not assumed to be intrinsically informative; instead, the detection objective determines how each discrepancy pattern should affect multimodal representation learning.

\begin{figure}[!t]
  \includegraphics[width=\columnwidth]{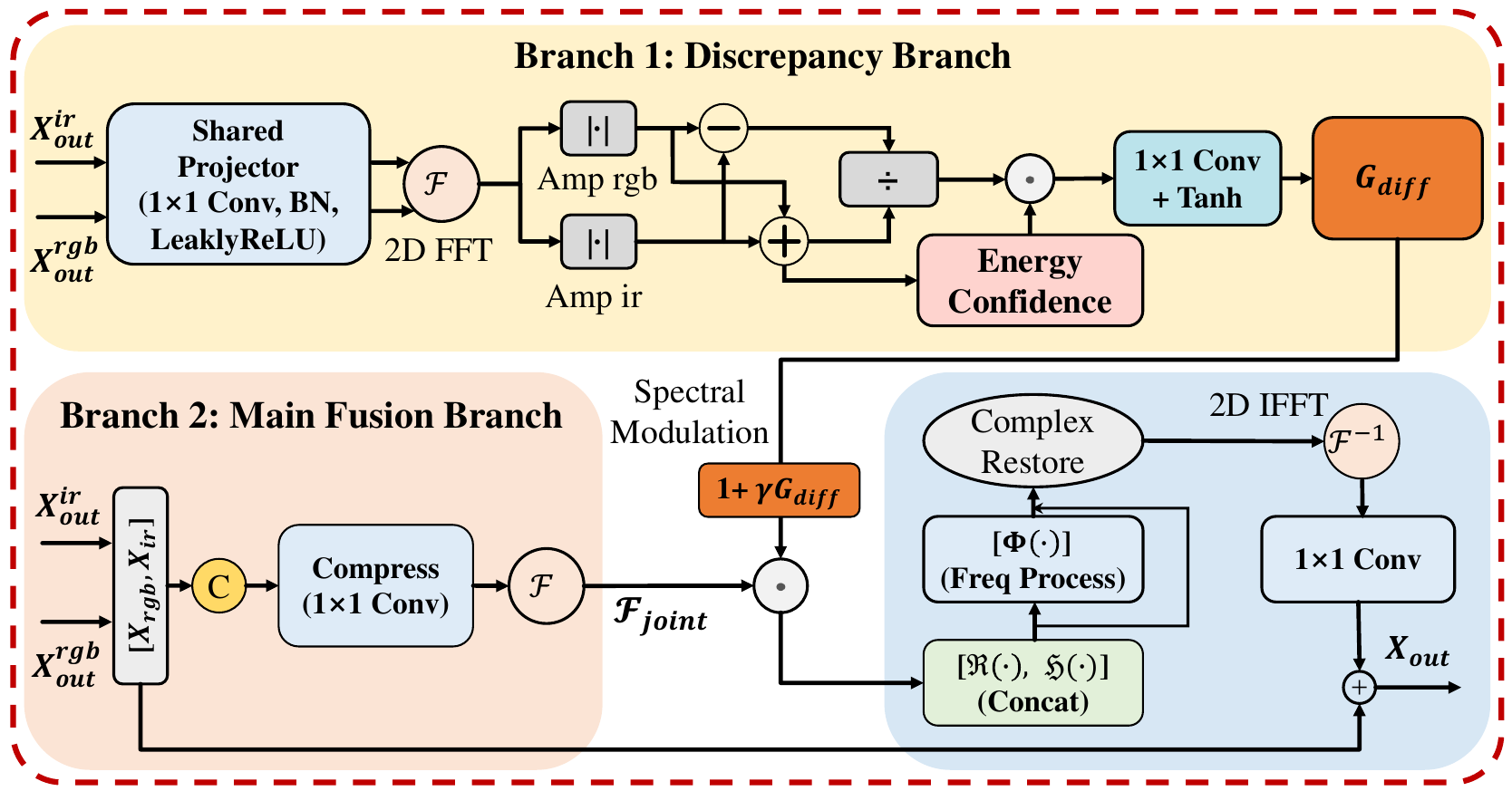}
  \caption{Overall architecture of the proposed DGSM}
  \label{fig2-DGSM}
\end{figure}

In parallel with the discrepancy branch, the original RGB and IR features are concatenated and projected into a compact joint representation:
\begin{equation}
X_{joint}
=
\mathcal{P}_{j}([X_{rgb},X_{ir}]),
\end{equation}
where $\mathcal{P}_{j}(\cdot)$ denotes a lightweight channel compression operator. Its joint spectrum is obtained as
\begin{equation}
Z_{joint}
=
\mathcal{F}(X_{joint}).
\end{equation}
We then use the signed discrepancy gate to perform bounded residual modulation:
\begin{equation}
Z_{mod}
=
Z_{joint}\odot
\left(1+\gamma G_{d}\right),
\label{eq:spectral_modulation}
\end{equation}
where $\gamma$ controls the maximum modulation magnitude and is empirically set to $0.5$. In this way, the original joint spectrum serves as a stable information basis, while cross-modal spectral discrepancy only controls an additional residual adjustment.

To further model interactions between the real and imaginary components, we convert the modulated complex spectrum into a real-valued representation:
\begin{equation}
Q=
[\operatorname{Re}(Z_{mod}),
 \operatorname{Im}(Z_{mod})].
\end{equation}
A lightweight spectral transformation $\Phi(\cdot)$ is then applied in a residual manner:
\begin{equation}
Q^{out}
=
Q+\Phi(Q).
\label{eq:fft_residual}
\end{equation}
The resulting real and imaginary components are recombined into a complex spectrum and transformed back to the spatial domain:
\begin{equation}
X_{fft}
=
\mathcal{F}^{-1}
\left(
\operatorname{Complex}
(Q^{out}_{real},Q^{out}_{imag})
\right).
\end{equation}
Finally, the reconstructed feature is projected back to the original multimodal dimensionality and combined with the concatenated input through a residual connection.

Overall, DGSM separates discrepancy estimation from multimodal content fusion: the discrepancy branch identifies where cross-modal spectral imbalance occurs, while the joint branch independently preserves the content to be fused. The former serves only as a learned control signal for the latter. This discrepancy-as-control formulation allows DGSM to exploit complementary spectral differences while reducing the risk of indiscriminately amplifying modality-specific noise.
\section{Experiments}
\label{sec4:Experiments}
\subsection{Datasets and Evaluation Metrics}
\label{sec4:Datasets}
\textbf{Datasets.}
We evaluate our model on three commonly used drone-based visible--infrared object detection datasets: DroneVehicle \cite{sun2022drone}, ESCVehicle \cite{song2026escvehicle}, and ATR-UMOD \cite{chen2025fusion}.

(1) DroneVehicle Dataset. The DroneVehicle dataset is a large-scale RGB--IR benchmark captured from aerial perspective, containing 953,087 vehicle instances across 28,439 RGB--IR image pairs. It covers diverse scenarios, including urban roads, residential areas, and parking lots, under both daytime and nighttime conditions. The dataset provides annotations for five vehicle categories: car, bus, truck, van, and freight car. Following the official data split, we use 17,990 image pairs for training, 1,469 for validation, and 8,980 for testing.

(2) ESCVehicle dataset. The ESCVehicle dataset is a drone-based visible-infrared benchmark containing 10,727 aligned RGB--IR image pairs with 369,714 annotated vehicle instances. It covers seven vehicle categories and diverse scenarios, including urban areas, parking lots, roads, tree-covered regions, rain, fog, snow, lakes, and hills. Following the official split, we use 6,983 image pairs for training, 507 for validation, and 3,237 for testing. The diverse scene annotations make ESCVehicle particularly suitable for evaluating the robustness of multimodal vehicle detection methods under challenging environmental conditions.

(3) ATR-UMOD Dataset. The ATR-UMOD dataset is a high-diversity UAV-based RGB--IR benchmark containing 13,353 aligned image pairs across 11 object categories. It covers diverse imaging conditions, including flight altitudes from 80 m to 300 m, camera angles from \(0^\circ\) to \(75^\circ\), all-day and all-year acquisition, as well as various weather, illumination, and scenario conditions. Each image pair is additionally annotated with six condition attributes, including altitude, angle, time, weather, illumination, and scenario, providing a comprehensive benchmark for evaluating multimodal detection under complex conditions. Following the official split, we use 11,850 image pairs for training and 1,503 for testing.

\textbf{Evaluation Metrics.}
We adopt standard evaluation metrics, including mAP$_{50}$ and mAP, to assess model performance on different datasets. Specifically, mAP$_{50}$ denotes the mean Average Precision at an IoU threshold of 0.50, while mAP is averaged over IoU thresholds from 0.50 to 0.95 with a step size of 0.05.

\subsection{Implementation Details}
\label{sec:Implementation}
All experiments are conducted on a single NVIDIA RTX 4090 GPU with 24 GB memory, using CUDA 12.1 and PyTorch 2.5.1. Our model is implemented by extending the Ultralytics \cite{yolo11_ultralytics} single-modal framework into a dual-stream multimodal architecture. To ensure a fair evaluation of the proposed architectural designs, all models are trained entirely from scratch without relying on any pre-trained weights. We train all models using the SGD optimizer with a momentum of 0.937, a weight decay of 0.0005, and an initial learning rate of 0.01. For DroneVehicle, ESCVehicle, and ATR-UMOD, the batch size is set to 16, and the number of training epochs is set to 150. Unless otherwise specified, the loss functions, other hyperparameters, and data augmentation settings follow the default configuration.

\begin{table*}[!t]
    \centering
    \caption{Comparison of FoCal with other methods on the DroneVehicle test set. The terms “RGB” and “IR” denote detection outcomes from visible images and infrared images, respectively, while “RGB+IR” reflects combined detection outcomes from fused infrared and visible images. The best and second-best results in each category are highlighted in \textcolor{red}{red} and \textcolor{blue}{blue}, respectively. $^\dagger$ denotes the results reproduced by ourselves under the same training and evaluation settings.}
    \label{tab:dronevehicle_comparison}
    \setlength{\tabcolsep}{2mm}
    \begin{tabular}{lcccc|ccccc|c}
    \toprule
    Methods & Pub. & Modality & mAP$_{50}$ & mAP & Car & Truck & Bus & Van & Freight car & Param\# \\
    \midrule
    S$^{2}$ANet \cite{han2021align}         & TGRS'21       & RGB         & 61.0 & 31.4   & 80.0 & 54.2 & 84.9 & 43.8 & 42.2   & 70.7M \\
    YOLO11n $^\dagger$ \cite{yolo11_ultralytics}  & Ultralytics'24 & RGB         & 69.1 & 47.9   & 93.6 & 62.0 & 90.4 & 50.4 & 49.2   & \textcolor{red}{2.7M} \\
    \midrule
    S$^{2}$ANet \cite{han2021align}         & TGRS'21       & IR          & 67.5 & 40.4    & 89.9 & 54.5 & 88.9 & 48.4 & 55.8   & 70.7M \\
    YOLO11n $^\dagger$ \cite{yolo11_ultralytics}  & Ultralytics'24 & IR          & 79.2 & 63.3    & 98.3 & 76.8 & 95.0 & 60.4  & 65.7   & \textcolor{red}{2.7M} \\
    \midrule
    YOLO11n-Dual $^\dagger$ \cite{yolo11_ultralytics}   & Ultralytics'24 & RGB+IR & 80.8 & 65.2   & \textcolor{blue}{98.5} & 80.5 & 95.8 & 62.2 & 67.0   & 4.0M \\
    C$^2$Former \cite{yuan2024c2former}      & TGRS'24       & RGB+IR & 74.2 & 47.5   & 90.2 & 68.3 & 89.8 & 58.5 & 64.4   & 118.5M \\
    CALNet \cite{he2023multispectral}        & ACM MM'23     & RGB+IR & 75.4 & 48.2   & 90.3 & 76.2 & 89.1 & 58.5 & 63.0   & 39.5M  \\
    OAFA \cite{chen2024weakly}               & CVPR'24       & RGB+IR & 77.1 & 50.1   & 90.1 & 75.4 & 89.8 & 61.8 & 68.2   &   -     \\
    M$^{2}$D-LIF \cite{zhao2025rethinking}   & ICCV'25       & RGB+IR & 81.4 & \textcolor{blue}{68.1} & 97.8 & 81.0 & \textcolor{blue}{96.0} & \textcolor{blue}{64.6} & 67.9 & 37.1M \\
    FusionMamba \cite{dong2025fusion}        & TMM'25        & RGB+IR & 79.2 & 56.0   & 96.7 & 80.2 & 95.3 & 64.0 & 59.8   & 287.6M \\
    WaveMamba \cite{zhu2025wavemamba}        & ICCV'25       & RGB+IR & 79.8 & 60.5   & 95.0 & 80.4 & 90.6 & 64.5 & 68.5   & 69.1M  \\
    C$^{2}$DFF-Net $^\dagger$ \cite{11180153}           & TGRS'25       & RGB+IR & \textcolor{blue}{82.0} & 66.7 & \textcolor{red}{98.6} & \textcolor{blue}{82.2} & 95.9 & 64.0 & \textcolor{blue}{69.5} & 6.6M \\
    COMO $^\dagger$ \cite{liu2026cross}                 & Inf. Fusion'26 & RGB+IR & 81.2 & 66.7   & 97.8 & 81.3 & 94.8 & 63.9 & 68.4   & 6.5M  \\
    CrossWeaver $^\dagger$ \cite{Yang_2026_CVPR}      & CVPR'26       & RGB+IR & 80.6 & 57.3   & 98.4 & 79.2 & 95.0 & 63.2 & 67.3   & 5.4M  \\
    FoCal (Ours)                             & --            & RGB+IR & \textcolor{red}{83.5} & \textcolor{red}{68.6} & \textcolor{red}{98.6} & \textcolor{red}{83.6} & \textcolor{red}{96.2} & \textcolor{red}{66.9} & \textcolor{red}{72.0} & \textcolor{blue}{3.0M} \\
    \bottomrule
    \end{tabular}
\end{table*}

\subsection{Comparisons with State-of-the-Art Methods}
\label{sec:Comparison}
\textbf{Comparison on DroneVehicle.}
Table~\ref{tab:dronevehicle_comparison} reports the comparison with representative single-modal and RGB--IR object detectors on the DroneVehicle test set. FoCal achieves the best overall performance, reaching 83.5\% $\mathrm{mAP}_{50}$ and 68.6\% mAP. Specifically, it surpasses the second-best C$^{2}$DFF-Net by 1.5 percentage points in $\mathrm{mAP}_{50}$ and M$^{2}$D-LIF by 0.5 percentage points in mAP. Compared with the YOLO11n-Dual baseline, FoCal further improves $\mathrm{mAP}_{50}$ from 80.8\% to 83.5\% and mAP from 65.2\% to 68.6\%, corresponding to gains of 2.7 and 3.4 percentage points, respectively. This demonstrates the effectiveness of the proposed cross-modal interaction and spectral calibration beyond conventional dual-stream fusion. At the category level, FoCal achieves APs of 98.6\%, 83.6\%, 96.2\%, 66.9\%, and 72.0\% for car, truck, bus, van, and freight car, respectively. It ranks first on four of the five categories and ties for the best result on car. Particularly, FoCal improves the second-best AP by 1.4, 2.3, and 2.5 percentage points for truck, van, and freight car, respectively, indicating consistent improvements across object categories. FoCal also maintains high computational efficiency. With only 3.0M parameters, it has the smallest model size among the compared RGB--IR methods with reported parameter counts, while outperforming substantially larger models such as M$^{2}$D-LIF (37.1M), WaveMamba (69.1M), and FusionMamba (287.6M). Moreover, FoCal reduces the parameter count by 25\% relative to YOLO11n-Dual (4.0M) while achieving higher detection accuracy, demonstrating a favorable accuracy--complexity trade-off.
\begin{table*}[!t]
    \centering
    \caption{Comparison of FoCal with other methods on the ESCVehicle dataset.}
    \label{tab:escvehicle_comparison}
    \setlength{\tabcolsep}{4mm}
    \begin{tabular}{lccc|cc}
    \toprule
    Method & Modality & mAP$_{50}$ & mAP & Param\# & FLOPs \\
    \midrule
    RoI Transformer \cite{ding2019learning} & RGB & 37.2 & 19.3 & 87.2M & 148.7G \\
    S$^2$ANet \cite{han2021align}           & RGB & 35.5 & 17.5 & 70.7M & 120.6G \\
    Oriented R-CNN \cite{xie2021oriented}   & RGB & 39.3 & 20.6 & 73.2M & 134.8G \\
    KLD \cite{yang2021learning}             & RGB & 32.9 & 16.2 & 41.9M & 131.2G \\
    \midrule
    RoI Transformer \cite{ding2019learning} & IR & 22.0 & 10.5 & 87.2M & 148.7G \\
    S$^2$ANet \cite{han2021align}           & IR & 21.1 & 9.7 & 70.7M & 120.6G \\
    Oriented R-CNN \cite{xie2021oriented}   & IR & 24.3 & 11.4 & 73.2M & 134.8G \\
    OSIV-Net \cite{zhang2025omni}           & IR & 37.0 & 19.1 & 10.8M & 29.38G \\
    \midrule
    CFT \cite{qingyun2021cross}             & RGB+IR & 41.7 & 23.9 & 44.7M & 44.0G \\
    ICAFusion \cite{shen2024icafusion}      & RGB+IR & 45.0 & 30.3 & 24.5M & 39.2G \\
    CALNet \cite{he2023multispectral}       & RGB+IR & 48.8 & 31.8 & 39.5M & 52.9G \\
    C$^2$Former \cite{yuan2024c2former}     & RGB+IR & 48.7 & 30.8 & 36.8M & 80.3G \\
    CMA-Det \cite{song2024misaligned}       & RGB+IR & 47.6 & 27.9 & 28.9M & 39.8G \\
    C$^2$-VeD \cite{song2026escvehicle}     & RGB+IR 
        & \textcolor{blue}{52.4} 
        & \textcolor{blue}{38.6} 
        & 16.5M 
        & \textcolor{blue}{13.1G} \\
    COMO $^\dagger$ \cite{liu2026cross}     & RGB+IR & 49.5 & 37.8 & 6.5M & 25.5G \\
    CrossWeaver $^\dagger$ \cite{Yang_2026_CVPR} 
        & RGB+IR & 45.4 & 31.0 
        & \textcolor{blue}{5.4M} 
        & 19.5G \\
    FoCal (Ours)                            & RGB+IR 
        & \textcolor{red}{54.8} 
        & \textcolor{red}{41.7} 
        & \textcolor{red}{3.0M} 
        & \textcolor{red}{8.4G} \\
    \bottomrule
    \end{tabular}
\end{table*}
\begin{table*}[t]
\centering
\caption{Comparison of FoCal with other methods on the ATR-UMOD dataset. All methods perform localization and classification using oriented bounding box (OBB) heads. The categories, car, SUV, van, bus, freight car, truck, motorcycle, trailer, excavator, crane, and tank truck, are abbreviated as CR, SV, VN, BS, FC, TK, ME, TR, ER, CE, and TT, respectively.}
\label{tab:atrumod_comparison}
\setlength{\tabcolsep}{2mm}
\begin{tabular}{l|c|cccccccccccc}
\toprule
Detectors & Modality & CR & SV & VN &
BS & FC & TK & TT & TR &
CE & ER & ME & mAP$_{50}$ \\
\midrule

S$^2$ANet \cite{han2021align}
& \multirow{5}{*}{RGB}
& 34.2 & 44.9 & 45.9 & 69.2 & 24.4 & 37.4 & 5.6
& 22.5 & 49.5 & 31.2 & 25.6 & 35.5 \\

ReDet \cite{han2021redet}
&
& 36.9 & 52.5 & 51.6 & 74.8 & 33.5 & 48.1 & 16.7
& 40.7 & 61.4 & 36.8 & 32.9 & 41.1 \\

RoI Transformer \cite{ding2019learning}
&
& 37.2 & 53.3 & 51.9 & 71.5 & 30.1 & 46.8 & 18.2
& 36.3 & 58.9 & 38.3 & 25.3 & 42.5 \\

Oriented R-CNN \cite{xie2021oriented}
&
& 36.9 & 52.5 & 51.6 & 74.8 & 33.5 & 48.1 & 16.7
& 40.7 & 61.7 & 36.8 & 32.9 & 44.2 \\

YOLOv5s \cite{yolo11_ultralytics}
&
& 45.8 & 60.7 & 57.5 & 75.2 & 41.6
& \textcolor{red}{52.1}
& 18.2 & 42.3 & 68.7 & 47.5 & 47.4 & 50.7 \\

\midrule

S$^2$ANet \cite{han2021align}
& \multirow{5}{*}{IR}
& 50.2 & 35.9 & 31.8 & 59.9 & 35.5 & 24.3 & 31.4
& 16.0 & 10.8 & 1.0 & 32.0 & 29.9 \\

ReDet \cite{han2021redet}
&
& 57.4 & 42.6 & 38.8 & 70.4 & 42.3 & 31.5 & 52.0
& 15.9 & 33.7 & 8.7 & 23.1 & 37.9 \\

RoI Transformer \cite{ding2019learning}
&
& 54.6 & 41.8 & 38.7 & 64.0 & 43.1 & 33.6 & 61.0
& 23.4 & 32.8 & 7.0 & 23.4 & 38.5 \\

Oriented R-CNN \cite{xie2021oriented}
&
& 57.5 & 41.6 & 36.8 & 63.8 & 43.5 & 28.6 & 64.3
& 28.5 & 44.2 & 6.9 & 23.9 & 40.0 \\

YOLOv5s \cite{yolo11_ultralytics}
&
& 65.8 & 51.2 & 51.6 & 75.3 & 53.1 & 38.9
& \textcolor{blue}{83.3}
& 46.2 & 57.9 & 12.0 & 42.7 & 52.5 \\

\midrule

UA-CMDet \cite{sun2022drone}
& \multirow{10}{*}{RGB+IR}
& 50.9 & 43.3 & 47.9 & 75.8 & 51.4 & 44.5 & 42.8
& 40.1 & 54.8 & 39.6 & 23.2 & 46.8 \\

C$^2$Former \cite{yuan2024c2former}
&
& 60.5 & 53.3 & 51.6 & 81.6 & 46.1 & 44.7 & 46.6
& 29.3 & 56.3 & 36.8 & 40.0 & 49.7 \\

TINet \cite{zhang2023illumination}
&
& 60.2 & 51.4 & 54.4 & 74.5 & 50.2 & 46.0 & 44.6
& 39.7 & 59.0 & 47.5 & 27.0 & 50.4 \\

CALNet \cite{he2023multispectral}
&
& 71.9
& \textcolor{red}{65.5}
& \textcolor{red}{71.0}
& 78.4 & 53.6 & 51.2 & 37.7
& 35.3 & 56.3 & 31.9 & 38.6 & 53.8 \\

OAFA \cite{chen2024weakly}
&
& 70.4 & 59.6 & 63.1 & 81.5 & 60.1 & 47.5 & 80.1
& 32.4 & 59.0 & 33.0 & 50.1 & 57.9 \\

YOLOrs \cite{sharma2020yolors}
&
& 73.2 & 62.6 & 66.3 & 81.8 & 61.1 & 48.2 & 70.3
& 37.6 & 64.3 & 41.8 & 52.9 & 60.0 \\

CrossWeaver $^\dagger$ \cite{Yang_2026_CVPR}
&
& \textcolor{blue}{74.2}
& 59.8 & 62.3
& \textcolor{blue}{87.3}
& 61.8 & 48.6 & 81.6
& 43.5 & 67.5 & 34.9
& \textcolor{blue}{56.6}
& 61.7 \\

PCDF \cite{chen2025fusion}
&
& 70.8 & 60.6 & 65.4 & 84.3
& \textcolor{blue}{62.1}
& \textcolor{blue}{51.3}
& \textcolor{red}{86.1}
& 42.5 & 71.1
& \textcolor{blue}{49.0}
& 51.2 & 63.1 \\

COMO $^\dagger$ \cite{liu2026cross}
&
& \textcolor{red}{76.0}
& \textcolor{blue}{62.9}
& \textcolor{blue}{67.6}
& 87.1
& \textcolor{red}{63.8}
& 47.6 & 82.8
& \textcolor{blue}{47.6}
& \textcolor{blue}{75.4}
& 42.2
& \textcolor{red}{59.2}
& \textcolor{blue}{64.2} \\

FoCal (Ours)
&
& 68.3 & 60.7 & 64.6
& \textcolor{red}{91.1}
& 61.5 & 46.4 & 82.3
& \textcolor{red}{52.1}
& \textcolor{red}{78.3}
& \textcolor{red}{58.5}
& 47.0
& \textcolor{red}{64.6} \\

\bottomrule
\end{tabular}
\end{table*}

\textbf{Comparison on ESCVehicle.}
Table~\ref{tab:escvehicle_comparison} reports the comparison with representative single-modal and RGB--IR detectors on the ESCVehicle dataset. FoCal achieves the best overall performance, reaching 54.8\% $\mathrm{mAP}_{50}$ and 41.7\% mAP. Compared with the second-best C$^{2}$-VeD, FoCal improves $\mathrm{mAP}_{50}$ and mAP by 2.4 and 3.1 percentage points, respectively. It also consistently outperforms other representative RGB--IR methods, demonstrating the effectiveness of the proposed cross-modal interaction and spectral calibration. Notably, FoCal achieves these improvements with only 3.0M parameters and 8.4G FLOPs, both being the lowest among all compared methods. Compared with C$^{2}$-VeD, FoCal reduces the model size from 16.5M to 3.0M and the computational cost from 13.1G to 8.4G while achieving higher accuracy. These results demonstrate that FoCal simultaneously improves detection effectiveness and computational efficiency.

\textbf{Comparison on ATR-UMOD.}
Table~\ref{tab:atrumod_comparison} presents the comparison with representative single-modal and RGB--IR detectors on the ATR-UMOD dataset. FoCal achieves the best overall performance with an $\mathrm{mAP}_{50}$ of 64.6\%. It also consistently outperforms other representative multimodal methods, including OAFA, COMO, and CrossWeaver, demonstrating the effectiveness of the proposed frequency-dependent cross-modal interaction and discrepancy-guided spectral modulation. At the category level, FoCal achieves the best AP on bus, trailer, crane, and excavator, reaching 91.1\%, 52.1\%, 78.3\%, and 58.5\%, respectively. The pronounced gains on these categories indicate that FoCal can effectively exploit complementary RGB--IR information across objects with diverse appearance, contributing to its superior overall detection performance.
\subsection{Ablation Studies}
\label{tab:ablation}
To validate the effectiveness of the proposed components in FoCal, we conduct ablation studies on DroneVehicle to examine the contributions of the architectural redesign, FADC, and DGSM, followed by a comparison of frequency interaction strategies.

\textbf{Redundancy Reduction.}
Considering the structural redundancy of dual-stream feature extraction, we adopt a lightweight redundancy reduction (RR) strategy. Specifically, the modality-specific branches are retained up to P4, while a shared P5 representation is constructed from the fused P4 feature through parameter-free downsampling followed by high-level semantic encoding, thereby avoiding duplicated deep feature extraction. In addition, RR employs a compact two-scale prediction hierarchy on P4 and P5. The high-resolution P3 representation remains involved in bidirectional multi-scale aggregation and continuously supplies fine-grained spatial cues to subsequent features, while final predictions are concentrated on the semantically stronger P4 and P5 levels. This design decouples feature-scale aggregation from prediction-scale allocation and provides a compact detection architecture without discarding high-resolution information. As shown in Table~\ref{table:ablation_dronevehicle}, RR improves $\mathrm{mAP}_{50}$, $\mathrm{mAP}_{75}$, and mAP from 80.8\%, 76.6\%, and 65.2\% to 81.9\%, 78.5\%, and 67.3\%, respectively, while reducing the parameter count from 4.01M to 2.63M and FLOPs from 9.5G to 6.8G. This avoids duplicated high-level computation and redundant dense predictions, yielding a more compact architecture while preserving detection accuracy.

\textbf{Effectiveness of Individual Components.}
Introducing RR improves these metrics to 81.9\%, 78.5\%, and 67.3\%, respectively, while reducing the parameter count from 4.01M to 2.63M and FLOPs from 9.5G to 6.8G. This provides a more compact and effective starting point for subsequent cross-modal modeling. Building upon RR, FADC further improves mAP$_{50}$, mAP$_{75}$, and mAP to 82.9\%, 79.5\%, and 68.2\%, respectively, with only 0.12M additional parameters and 0.5G FLOPs. This verifies the effectiveness of modeling different cross-modal interaction patterns for low- and high-frequency components and transferring the resulting frequency cues back to the original feature streams for calibration. Independently adding DGSM to RR achieves 82.6\% mAP$_{50}$, 79.1\% mAP$_{75}$, and 68.0\% mAP. The consistent improvement demonstrates that confidence-weighted spectral discrepancy provides complementary guidance for regulating the joint multimodal spectrum. Combining FADC and DGSM yields the best overall performance, reaching 83.5\% mAP$_{50}$, 79.9\% mAP$_{75}$, and 68.6\% mAP. These results indicate that the two modules provide complementary benefits: FADC focuses on frequency-dependent cross-modal interaction and original-domain calibration, whereas DGSM regulates the joint spectral representation using cross-modal discrepancy cues. 

\begin{table}
    \centering
    \caption{Ablation study of individual components in FoCal on the DroneVehicle dataset.}
    \label{table:ablation_dronevehicle}
    \setlength{\tabcolsep}{1.3mm}
    \begin{tabular}{ccc|ccc|cc}
    \toprule
    RR & FADC & DGSM & mAP$_{50}$ &mAP$_{75}$ & mAP & Param\# & FLOPs \\
    \midrule
    \ding{55} & \ding{55}       & \ding{55}  & 80.8  & 76.6 & 65.2  & 4.01M  & 9.5G \\
    \ding{51} & \ding{55}       & \ding{55}  & 81.9  & 78.5 & 67.3  & 2.63M  & 6.8G \\
    \ding{51} & \ding{51}       & \ding{55}  & 82.9  & 79.5 & 68.2  & 2.75M  & 7.3G \\
    \ding{51} & \ding{55}       & \ding{51}  & 82.6  & 79.1 & 68.0  & 2.84M  & 7.9G \\
    \ding{51} & \ding{51}       & \ding{51}  & 83.5  & 79.9 & 68.6  & 2.96M  & 8.4G \\
    \bottomrule
    \end{tabular}
\end{table}

\begin{table}[!t]
\centering
\caption{Ablation study of different frequency interaction strategies on the DroneVehicle dataset. ``Ex.'' and ``Con.'' denote selective exchange and structural consensus, respectively.}
\label{tab:ablation_frequency_interaction_strategies}
\setlength{\tabcolsep}{1.5mm}
\begin{tabular}{lccccccc}
\toprule
Variant & HF & LF & mAP$_{50}$ & mAP$_{75}$ & mAP & Param\# & FLOPs \\
\midrule
Sym.    & Ex.  & Ex.  & 82.3 & 78.8 & 67.6 & 2.71M & 7.1G \\
Sym.    & Con. & Con. & 82.7 & 79.2 & 68.0 & 2.79M & 7.5G \\
Reverse & Con. & Ex.  & 82.4 & 78.9 & 67.7 & 2.75M & 7.3G \\
\textbf{Ours} & Ex. & Con. & \textbf{82.9} & \textbf{79.5} & \textbf{68.2} & 2.75M & 7.3G \\
\bottomrule
\end{tabular}
\end{table}

\begin{figure*}[!t]
\centering
  \includegraphics[width=.9\textwidth]{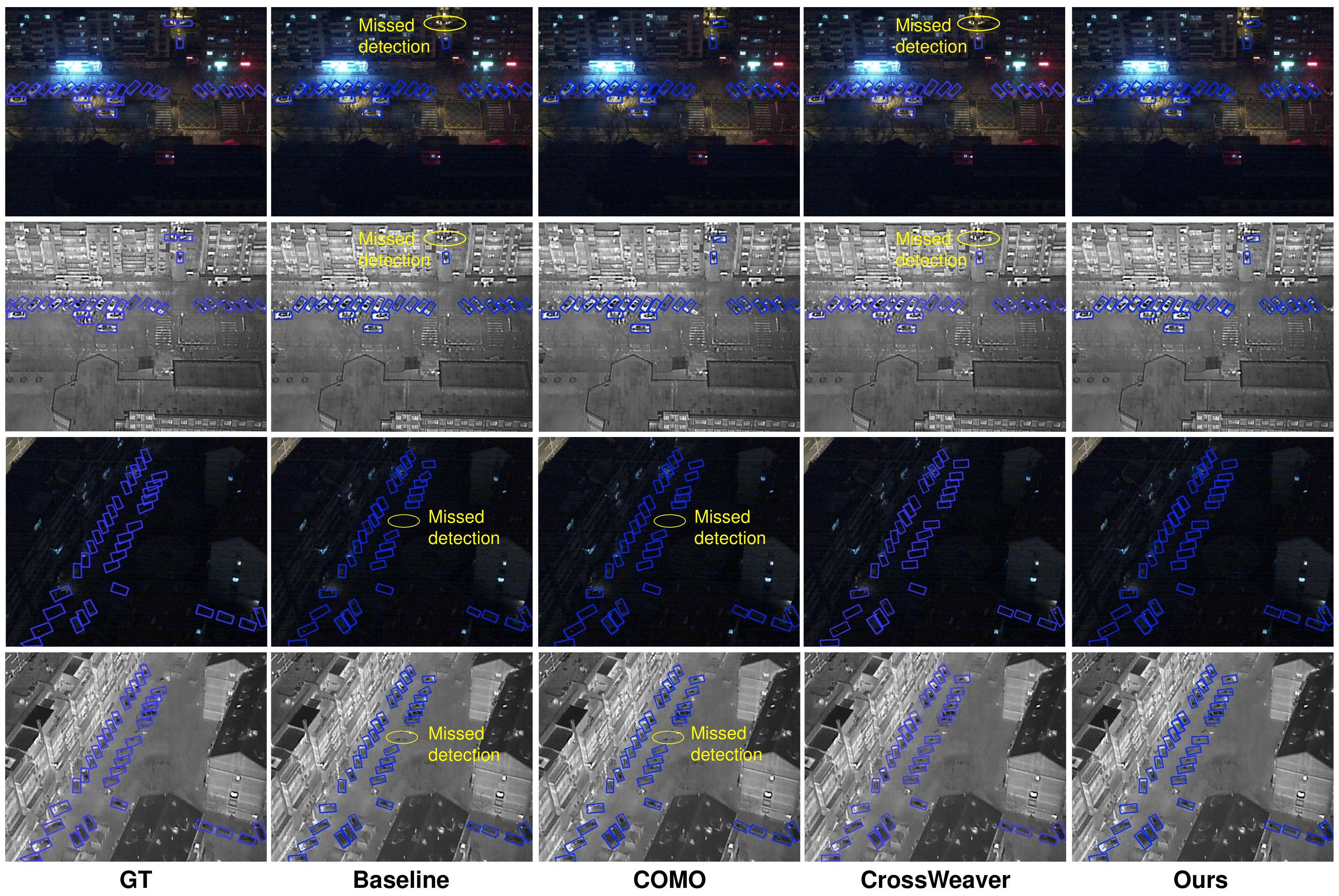}
  \caption{Qualitative comparison of the proposed method against other competing methods under complex aerial scenarios.}
  \label{fig-detect}
\end{figure*}

\begin{figure*}[!t]
\centering
  \includegraphics[width=.8\textwidth]{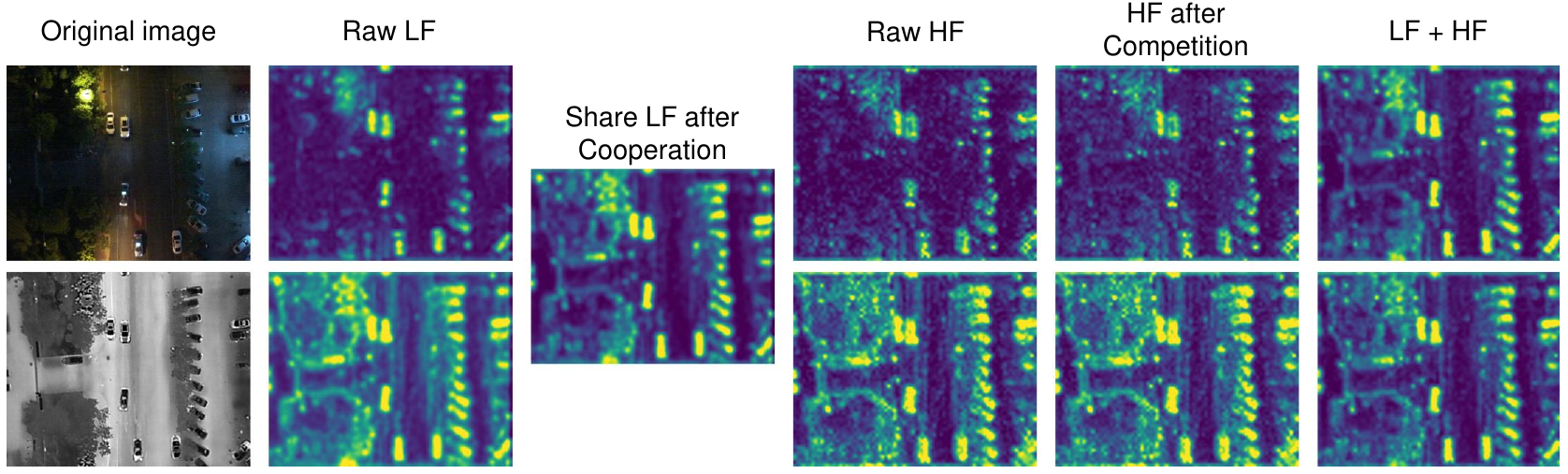}
  \caption{Visualization of the FADC frequency pathway at P3. The top and bottom rows correspond to RGB and IR. From left to right, the panels show the input images, raw low-frequency (LF) components, the shared LF representation after cooperative interaction, raw high-frequency (HF) components, HF components after competitive interaction, and the recomposed LF+HF features. One shared LF map is produced, whereas the HF and recomposed features remain modality-specific.}
  \label{fig-P3-FADC-Vis}
\end{figure*}

\textbf{Effectiveness of Frequency-Dependent Interaction.}
To investigate whether different frequency components favor different cross-modal interaction patterns, we compare four configurations in Table~\ref{tab:ablation_frequency_interaction_strategies}. Applying selective exchange to both frequency bands (\textit{Sym.Ex.}) achieves 82.3\% mAP$_{50}$, 78.8\% mAP$_{75}$, and 67.6\% mAP, while uniformly adopting structural consensus (\textit{Sym.Con.}) improves the results to 82.7\%, 79.2\%, and 68.0\%, respectively. These results show that both interaction mechanisms are effective, but a uniform strategy does not explicitly account for the distinct characteristics of different frequency components. We further reverse the proposed assignment by applying structural consensus to high-frequency components and selective exchange to low-frequency components (\textit{Reverse}). Under the same model complexity as our design, this configuration obtains 82.4\% mAP$_{50}$, 78.9\% mAP$_{75}$, and 67.7\% mAP. In contrast, our frequency-dependent strategy applies selective exchange to high-frequency components and structural consensus to low-frequency components, improving the three metrics to 82.9\%, 79.5\%, and 68.2\%, respectively. These results support the proposed frequency-dependent interaction principle: low-frequency components benefit from cross-modal structural consensus, whereas high-frequency components are better modeled by preserving modality-specific information through selective exchange.
\begin{figure}[!t]
  \includegraphics[width=\columnwidth]{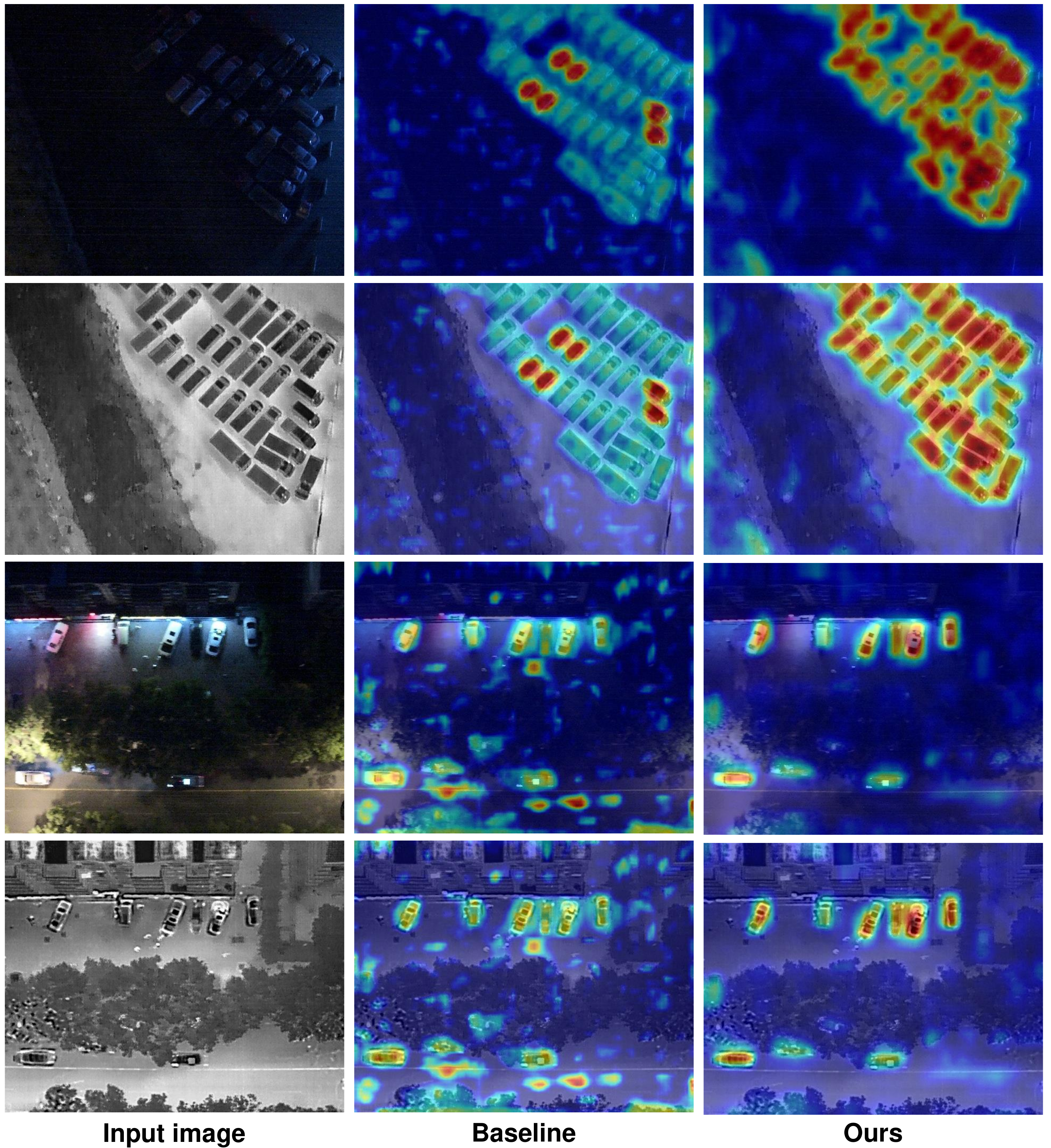}
  \caption{Comparison of heatmap visualizations across different input modalities. From top to bottom: visible and infrared modalities. From left to right: baseline model and the proposed FoCal.}
  \label{fig-heatmap}
\end{figure}

\subsection{Visualization}
\label{subsec_Visualization}
We further provide qualitative visualizations to examine the behavior of FoCal in challenging aerial scenes. As shown in Fig.~\ref{fig-detect}, the compared examples contain densely distributed vehicles under low illumination and weak visual contrast. In the first scene, both the baseline and CrossWeaver miss the highlighted target, whereas COMO and FoCal successfully detect it. In the second scene, the baseline and COMO fail to identify the highlighted vehicle within the densely arranged parking area, while CrossWeaver and FoCal recover it. In both cases, FoCal successfully preserves the target responses while maintaining stable detections for surrounding vehicles. The paired RGB and IR observations further illustrate the complementary characteristics of the two modalities, especially when target visibility is degraded in the visible spectrum. These results qualitatively demonstrate that FoCal can more effectively exploit complementary cross-modal cues in complex aerial environments.

Fig.~\ref{fig-P3-FADC-Vis} visualizes the intermediate frequency representations produced by FADC. The RGB and IR low-frequency maps exhibit different response distributions before interaction, while the consensus operation progressively consolidates them into a shared representation with clearer structural responses around vehicle regions. In contrast, the high-frequency branch maintains two modality-specific streams after selective exchange, preserving distinct local details while incorporating complementary cues from the counterpart modality. The response heatmaps in Fig.~\ref{fig-heatmap} further illustrate the effect of this interaction on the final representation. Compared with the baseline, FoCal produces more concentrated activations around vehicle locations and weaker diffuse responses in background regions, particularly under low-contrast RGB conditions. The IR branch also maintains stable responses on target regions, indicating that complementary thermal cues are effectively retained during fusion.

\begin{table}[!t]
\centering
\caption{Comparison of model complexity and inference efficiency. Latency and FPS are measured using the trained \texttt{.pt} models in PyTorch on an NVIDIA RTX 4090 with a batch size of 1.}
\label{tab:inference-efficiency}
\setlength{\tabcolsep}{1mm}
\begin{tabular}{lccccc}
\toprule
Method & Param\# & FLOPs  & mAP$_{50}$ & Latency (ms/pair) $\downarrow$ & FPS $\uparrow$ \\
\midrule
CrossWeaver \cite{Yang_2026_CVPR}     & 5.4M & 19.5G  & 80.6 & 32.1 & 31.2 \\
COMO \cite{liu2026cross}              & 6.5M & 25.5G  & 81.2 & 15.5 & 64.5 \\
\textbf{FoCal (Ours)}                 & 3.0M & 8.4G   & \textbf{83.5} & \textbf{8.8} & \textbf{113.6} \\
\bottomrule
\end{tabular}
\end{table}

\section{Discussion}
\label{Discussion}
Overall, the experimental results on DroneVehicle, ESCVehicle, and ATR-UMOD demonstrate that FoCal consistently achieves strong detection performance across different aerial RGB--IR benchmarks. In particular, the improvements in both overall detection accuracy and category-level performance indicate that frequency-dependent cross-modal interaction can effectively exploit complementary information between visible and infrared modalities. Meanwhile, the compact architecture of FoCal suggests that these accuracy gains do not rely on increasing model capacity, providing a favorable basis for practical multimodal detection.

Beyond detection accuracy, computational efficiency is particularly important for aerial platforms, where real-time processing and limited computing resources are common constraints. As shown in Table~\ref{tab:inference-efficiency}, FoCal achieves 83.5\% mAP$_{50}$ with only 3.0M parameters and 8.4G FLOPs, while reaching 113.6 FPS with a latency of 8.8 ms per RGB--IR pair on an NVIDIA RTX 4090. In comparison, COMO achieves 81.2\% mAP$_{50}$ at 64.5 FPS, while CrossWeaver obtains 80.6\% mAP$_{50}$ at 31.2 FPS. The lower FPS of COMO and CrossWeaver may be associated with their more elaborate cross-modal interaction pipelines, including cross-Mamba-based global--local modeling in COMO and deformation-aware hierarchical interaction in CrossWeaver. In contrast, FoCal adopts a more compact frequency-oriented interaction design, reducing inference overhead while preserving strong detection accuracy. Thus, FoCal simultaneously provides higher detection accuracy, lower computational complexity, and substantially higher forward-pass throughput. These characteristics indicate its strong potential for real-time RGB--IR object detection and make it particularly suitable for future deployment on aerial and other latency-sensitive multimodal perception platforms. Future work will therefore investigate end-to-end deployment optimization, including mixed-precision inference, TensorRT acceleration, and hardware-aware optimization on embedded and edge computing platforms.
\section{Conclusion}
In this work, we proposed \textbf{FoCal}, a frequency-oriented framework for aerial RGB--IR object detection. FoCal explicitly exploits the distinct characteristics of different frequency components to improve cross-modal interaction. Specifically, FADC establishes structural consensus in the low-frequency branch while preserving modality-specific information through selective high-frequency exchange, and further transfers the resulting frequency-aware cues to calibrate the original feature streams. DGSM complements this process by converting confidence-weighted spectral discrepancy into a signed modulation signal for adaptive regulation of the joint multimodal spectrum. Extensive experiments on DroneVehicle, ESCVehicle, and ATR-UMOD demonstrate that FoCal consistently achieves state-of-the-art performance, with $\mathrm{mAP}_{50}$ values of 83.5\%, 54.8\%, and 64.6\%, respectively. Meanwhile, the model contains only 3.0M parameters and exhibits favorable inference efficiency, highlighting a strong trade-off between detection accuracy and computational cost. Future work will explore hardware-aware optimization and extend frequency-dependent interaction to temporal and more diverse multimodal perception scenarios. 



 
%

\bibliography{References}
\bibliographystyle{ieeetr}

\end{document}